\documentclass[11pt,a4paper]{article}
\usepackage[hyperref]{naaclhlt2018}
\usepackage{times}
\usepackage{latexsym}

\usepackage{amsmath}
\usepackage{amssymb}
\usepackage{multirow}
\usepackage[linesnumbered,lined,commentsnumbered]{algorithm2e}

\aclfinalcopy 

\usepackage{color}
\usepackage{soul}

\definecolor{DarkRed}{RGB}{130,25,0}
\sethlcolor{yellow}

\newenvironment{termsetenv}{$\{$}{$\}$}
\newcommand{\termset}[1]{\begin{termsetenv}\term{#1}\end{termsetenv}}
\newcommand{\term}[1]{\textit{#1}}
\newcommand{\catname}[1]{\textsc{\small #1}}
\newcommand{\featurename}[1]{\textsc{#1}}
\newcommand{\analogy}[4]{\term{#1}:\term{#2}::\term{#3}:\term{#4}}

\newcommand{\R}{\mathbb{R}}

\DeclareMathOperator{\pmi}{PMI}
\DeclareMathOperator{\ppmi}{PPMI}
\DeclareMathOperator{\apmi}{APMI}
\DeclareMathOperator{\appmi}{APPMI}

\DeclareMathOperator{\simmatrix}{SimMatrix}

\DeclareMathOperator{\precision}{Prec}
\DeclareMathOperator{\map}{MAP}

\newcommand{\mitof}{M^{\mathcal{V}\rightarrow\mathcal{C}}}
\newcommand{\mftoi}{M^{\mathcal{C}\rightarrow\mathcal{V}}}
\newcommand{\ditof}{D^{\mathcal{V}\rightarrow\mathcal{C}}}
\newcommand{\dftoi}{D^{\mathcal{C}\rightarrow\mathcal{V}}}

\title{Robust Handling of Polysemy via Sparse Representations}

\author{Abhijit A. Mahabal \\
  Google \\
  {\tt \small amahabal@google.com} \\\And
  Dan Roth \\
  University of Pennsylvania \\
  {\tt \small danroth@seas.upenn.edu} \\\And
  Sid Mittal \\
  Google \\
  {\tt \small sidmittal@google.com} \\  
  }

\date{}

\begin{document}
\maketitle

\begin{abstract}
Words are polysemous and multi-faceted, with many shades of meanings. We suggest that sparse distributed representations are more suitable than other, commonly used, (dense) representations to express these multiple facets, and present {\em Category Builder}, a working system that, as we show, makes use of sparse representations to support multi-faceted lexical representations. We argue that the set expansion task is well suited to study these meaning distinctions since a word may belong to multiple sets with a different reason for membership in each. We therefore exhibit the performance of {\em Category Builder} on this task, while showing that our representation captures at the same time {\em analogy} problems such as ``the Ganga of Egypt" or ``the Voldemort of Tolkien". {\em Category Builder} is shown to be a more expressive lexical representation and to outperform dense representations such as Word2Vec in some analogy classes despite being shown only two of the three input terms. 
\end{abstract}

\section{Introduction}
Word embeddings have received much attention lately because of their ability to represent similar words as nearby points in a vector space, thus supporting better generalization when comparisons of lexical items are needed, and boosting the robustness achieved by some deep-learning systems.
However, a given surface form often has multiple meanings, complicating this simple picture. \newcite{arora2016linear} showed that the vector corresponding to a polysemous term 
often is not close to any of that of its individual senses, thereby breaking the similar-items-map-to-nearby-points promise. The polysemy wrinkle is not merely an irritation but, in the words of \newcite{pustejovsky1997lexical}, ``one of the most intractable problems for language processing studies".

Our notion of Polysemy here is quite broad, since words can be similar to one another along a variety of dimensions. The following three pairs each has two similar items: 
(a) \termset{ring, necklace}, (b) \termset{ring, gang}, and (c) \termset{ring, beep}. Note that “ring” is similar to all words that appear as second words in these pairs, but for different reasons, {\em defined by the second token} in the pairs. While this example used different senses of \term{ring}, it is easy to find examples where a single sense has multiple {\em facets}: \term{Clint Eastwood}, who is both an actor and a director, shares different aspects with directors than with actors, and \term{Google}, both a website and a major corporation, is similar to \term{Wikipedia} and \term{General Electric} along different dimensions. 

Similarity has typically been studied pairwise: that is, by asking how similar item \term{A} is to item \term{B}. A simple modification sharply brings to fore the issues of facets and polysemy. This modification is best viewed through the task of \emph{set expansion}~\cite{ wang2007language,davidov2007fully,jindal2011learning}, which highlights the similarity of an item (a candidate in the expansion) to a set of “seeds” in the list. Given a few seeds (say, \termset{Ford, Nixon}), what else belongs in the set? Note how this expansion is quite different from the expansion of \termset{Ford, Chevy}, and the difference is one of \textit{Similar How}, since whether a word (say, \term{BMW} or \term{FDR}) belongs in the expansion depends not just on how much commonality it shares with \term{Ford} but on {\em what} commonality it shares. Consequently, this task allows the same surface form to belong to multiple sets, by virtue of being similar to items in distinct sets {\em for different reasons}. The facets along which items are similar is implicitly defined by the members in the set.

In this paper, we propose a context sensitive version of similarity based on highlighting shared facets.  We do this by developing a {\em sparse representation} of words that simultaneously captures all facets of a given surface form. This allows us to define a notion of contextual similarity, in which \term{Ford} is similar to \term{Chevy} (e.g., when \term{Audi} or \term{BMW} is in the context) but similar to \term{Obama} when \term{Bush} or \term{Nixon} is in the context (i.e., in the seed list). In fact, it can even support multi-granular similarity since while \termset{Chevy, Chrysler, Ford} represent the facet of \catname{American cars}, \termset{Chevy, Audi, Ford} define that of \catname{cars}. Our contextual similarity is better able to mold itself to this variety since it moves away from the  one-size-fits-all nature of cosine similarity. 

We exhibit the strength of the representation and the 
contextual similarity metric we develop by comparing its performance on both set expansion and analogy problems with dense representations.

\section{Senses and Facets}


The present work does not attempt to resolve the Word Sense Disambiguation (WSD) problem. Rather, our goal is to advance a lexical representation and a corresponding context sensitive similarity metric that, together, get around explicitly solving WSD.

Polysemy is intimately tied to the well-explored field of WSD so 
it is natural to expect techniques from WSD to be relevant. If WSD could neatly separate senses, the set expansion problem could be approached thus. \term{Ford} would split into, say, two senses: \term{Ford-1} for the car, and \term{Ford-2} for the president, and expanding \termset{Ford, Nixon} could be translated to expanding \termset{Ford-2, Nixon}. 
Such a representational approach is taken by many authors when they embed the different senses of words as distinct points in an embedding space \cite{reisinger2010multi, huang2012improving,neelakantan2014efficient,li2015multi}.

Such approaches run into what we term \emph{the Fixed Inventory Problem.} Either senses are obtained from a hand curated resource such as a dictionary, or are induced from the corpus directly by mapping contexts clusters to different senses. In either case, however, by the time the final representation (e.g., the embedding) is obtained, the number of different senses of each term has become fixed: all decisions have been made relating to how finely or coarsely to split senses.

How to split senses is a hard problem: dictionaries such as NOAD list coarse senses and split these further into fine senses, and it is unclear what granularity to use: should each fine sense correspond to a point in the vector space, or should, instead, each coarse sense map to a point? Many authors \cite[for example]{hofstadter2013surfaces} discuss how the various dictionary senses of a term are not independent. Further, if context clusters map to senses, the word \term{whale}, which is seen both in mammal-like contexts (e.g., ``whales have hair") and water-animal contexts (``whales swim"), could get split into separate points. Thus, the different \emph{senses} that terms are split into may instead be distinct \emph{facets}. This is not an idle theoretical worry: such facet-based splitting is evident in \newcite[Table 3]{neelakantan2014efficient}. Similarly, in the vectors they released, \term{november} splits into ten senses, likely based on facets. Once split, for subsequent processing, the points are independent.

In contrast to such explicit, prior, splitting, 
in the Category Builder approach developed here, relevant contexts are chosen given the task at hand, and if multiple facets are relevant (as happens, for example, in \termset{whale, dolphin, seal}, whose expansion should rank aquatic mammals highest), all these facets influence the expansion; if only one facet is of relevance (as happens in \termset{whale, shark, seahorse}), the irrelevant facets get ignored.

\section{Related Work}

In this section, we situate our approach within the relevant research landscape. Both \term{Set Expansion} and \term{Analogies} have a long history, and both depend on \term{Similarity}, with an even longer history.

\subsection{Set Expansion}

Set Expansion is the well studied problem of expanding a given set of terms by finding other semantically related terms. Solutions fall into two large families, differing on whether the expansion is based on a preprocessed, limited corpus \cite[for example]{shen2017setexpan} or whether a much larger corpus (such as the entire web) is accessed on demand by making use of a search engine such as Google \cite[for example]{wang2007language}.

Each family has its advantages and disadvantages. ``Open web" techniques that piggyback on Google can have coverage deep into the tail. These typically rely on some form of Wrapper Induction, and tend to work better for sets whose instances show up in lists or other repeated structure on the web, and thus perform much better on sets of nouns than on sets of verbs or adjectives. By contrast, ``packaged" techniques that work off a preprocessed corpus are faster (no Google lookup needed) and can work well for any part of speech, but are of course limited to the corpus used. These typically use some form of distributional similarity, which can compute similarity between items that have never been seen together in the same document; approaches based on shared memberships in lists would need a sequence of overlapping lists to achieve this.   Our work is in the ``packaged" family, and we use sparse representations used for distributional similarity.

\newcite{gyllensten2018distributional} compares two subfamilies within the packaged family: \emph{centrality}-based methods use a prototype of the seeds (say, the centroid) as a proxy for the entire seed set and \emph{classification}-based methods (a strict superset), which produce a classifier by using the seeds. Our approach is classification-based.

It is our goal to be able to expand nuanced categories. For example, we want our solution to expand the set \termset{pluto, mickey}---both Disney characters---to other Disney characters. That is, the context {\em mickey} should determine what is considered `similar' to pluto, rather than being biased by the more dominant sense of \term{pluto}, to determine that \term{neptune} is similar to it.  Earlier approaches such as \newcite{rong2016egoset} approach this problem differently: they expand to both planets and Disney characters, and then attempt to cluster the expansion into meaningful clusters. 

\subsection{Analogies}
Solving analogy problems usually refers to proportional analogies, such as \analogy{hand}{glove}{foot}{?}.
\newcite{mikolov2013linguistic} showed how word embeddings such as Word2Vec capture linguistic regularities and thereby solve this. \newcite{turney2012domain} used a pair of similarity functions (one for \emph{function} and one for \emph{domain}) to address the same problem.

There is a sense, however, that the problem is \emph{overdetermined}: in many such problems, people can solve it even if the first term is not shown. That is, people easily answer ``What is the \term{glove} for the \term{foot}?". People also answer questions such as ``What is the Ganga of Egypt?" without first having to figure out the unprovided term \term{India} (or is the missing term \term{Asia}? It doesn't matter.) \newcite{hofstadter2013surfaces} discuss how our ability to do these analogies is central to cognition.

The current work aims to tackle these \emph{non-proportional} analogies and in fact performs better than Word2Vec on some analogy classes used by \newcite{mikolov2013linguistic}, despite being shown one fewer term.

The approach is rather close to that used by \newcite{turney2012domain} for a different problem: \emph{word compounds}. Understanding what a \term{dog house} is can be phrased as ``What is the house of a dog?", with \term{kennel} being the correct answer. This is solved using the pair of similarity functions mentioned above. The evaluations provided in that paper are for \emph{ranking}: which of five provided terms is a match. Here, we apply it to non-proportional analogies and evaluate for retrieval, where we are ranking over all words, a significantly more challenging problem.

To our knowledge, no one has presented a computational model for analogies where only two terms are provided. We note, however, that \protect\newcite{linzen2016issues} briefly discusses this problem.

\subsection{Similarity}

Both Set Expansion and Analogies depend on a notion of similarity. Set Expansion can be seen as finding items most similar to a category, and Analogies can be seen as directly dependent on similarities (e.g., in the work of \newcite{turney2012domain}).

Most current approaches, such as word embeddings, produce a context independent similarity. In such an approach, the similarity between, say, \term{king} and \term{twin} is some fixed value (such as their cosine similarity). However, depending on whether we are talking about bed sizes, these two items are either closely related or completely unrelated, and thus context dependent.

Psychologists and Philosophers of Language have long pointed out that similarity is subtle. It is sensitive to context and subject to priming effects. Even the very act of categorization can change the perceived similarity between items \cite{goldstone2001altering}. \newcite[p. 275]{medin1993respects} tell a story, from the experimental psychology trenches, that supports representation morphing when they conclude that ``the effective representations of constituents are determined in the context of the comparison, not prior to it".

Here we present a malleable notion of similarity that can adapt to the wide range of human categories, some of which are based on narrow, superficial similarities (e.g., \catname{blue things}) while others share family resemblances (\`a la Wittgenstein). Even in a small domain such as movies, in different contexts, similarity may be driven by who the director is, or the cast, or the awards won. Furthermore, to the extent that the contexts we use are human readable, we also have a mechanism for explaining what makes the terms similar.

There is a lot of work on the context-dependence of human categories and similarities in Philosophy, in Cognitive Anthropology and in Experimental Psychology \cite[for example, survey this space from various theoretical standpoints]{lakoff1987women,ellis1993language,agar1994language,goldstone2001altering,hofstadter2013surfaces}, but there are not, to our knowledge, unsupervised computational models of these phenomena. 

\section{Representations and Algorithms}

This section describes the representation and corresponding algorithms that perform set expansion in Category Builder (CB).

\subsection{Sparse Representations for Expansion}\label{technical}
We use the traditional word representation that distributional similarity uses \cite{turney2010frequency}, and that is commonly used in fields such as context sensitive spelling correction and grammatical correction \cite{GoldingRo99,RozovskayaRo14}; namely,words are  associated with some ngrams that capture the contexts in which they occur -- all contexts are represented in a sparse vector corresponding to a word. Following \newcite{levy2014linguistic}, we call this representation \textit{explicit}.

\textbf{Generating Representations.} We start with web pages and extract words and phrases from these, as well as the contexts they appear in. An aggregation step then calculates the strengths of word to context and context to word associations.

\textbf{Vocabulary.} The vocabulary is made up of words (nouns, verbs, adjectives, and adverbs) and some multi-word phrases. To go beyond words, we use a named entity recognizer to find multi-word phrases such as \term{New York}. We also use one heuristic rule to add certain phrasal verbs (e.g., \term{take shower}), when a verb is directly followed by its direct object. We lowercase all phrases, and drop those phrases seen infrequently. The set of all words is called the vocabulary, $\mathcal{V}$.

\textbf{Contexts.} Many kinds of contexts have been used in literature. \newcite{levy2018word} provides a comprehensive overview. We use contexts derived from syntactic parse trees using about a dozen heuristic rules. For instance, one rule deals with nouns modified by an adjective, say, \term{red} followed by \term{car}. Here, one of the contexts of \term{car} is \featurename{ModifiedBy\#red}, and one of the contexts of \term{red} is \featurename{Modifies\#car}. Two more examples of contexts: \featurename{ObjectOf\#eat} and \featurename{SubjectOf\#write}. The set of all contexts is denoted $\mathcal{C}$.

\textbf{The Two Vocabulary$\Leftrightarrow$Context matrices.} For vocabulary $\mathcal{V}$ and contexts $C$, we produce \emph{two} matrices, $\mitof$ and $\mftoi$. Many measures of association between a word and a context have been explored in the literature, usually based on some variant of \textit{pointwise mutual information}.

PPMI (\emph{Positive} PMI) is the typically used measure. If $P(w)$, $P(c)$ and $P(w, c)$ respectively represent the probabilities that a word is seen, a context is seen and the word is seen in that context, then 
\begin{equation}
\pmi(w, c)=\log\frac{P(w,c)}{P(w)P(c)}
\end{equation}
\begin{equation}\label{eqn:ppmi}\ppmi(w, c) = max(0, \pmi(w, c))\end{equation}

PPMI is widely used, but comments are in order regarding the ad-hocness of the ``0" in Equation \ref{eqn:ppmi}. There is seemingly a good reason to choose 0 as a threshold: if a word is seen in a context more than by chance, the PMI is positive, and a 0 threshold seems sensible. However, in the presence of polysemy, especially lopsided polysemy such as \term{Cancer} (disease and star sign), a ``0" threshold is arbitrary: even if every single occurrence of the star sign sense of \term{cancer} was seen in some context $c$ (thereby crying out for a high PMI), because of the rarity of that sense, the overall PMI between $c$ and (non-disambiguated) \term{Cancer} may well be negative. Relatedly, Shifted PPMI \cite{levy2014neural} uses a non-0 cutoff.

Another well known problem with PPMI is its large value when the word or the context is rare, and even a single occurrence of a word-context pair can bloat the PMI (see \citealp{role2011handling}, for fixes that have been proposed). We introduce a 
new variant we call \emph{Asymmetric PMI}, which takes frequency into account by adding a second log term, and is asymmetric because in general $P(w|c)\ne P(c|w)$:
\begin{equation}\label{eqn:apmi}
\begin{split}
\apmi(w, c) &= \pmi(w, c) + \log\frac{P(w, c)}{P(w)}\\
&= \log\frac{P(w, c)^2}{P(w)^2P(c)}
\end{split}
\end{equation}

This is asymmetric because $\apmi(c, w)$ has $P(c)$ in the denominator of the extra log term.

What benefit does this modification to PMI provide? Consider a word and two associated contexts, $c_1$ and $c_2$, where the second context is significantly rarer. Further, imagine that the PMI of the word with either feature is the same. The word would have been seen in the rarer context only a few times, and this is more likely to have been a statistical fluke. In this case, the APMI with the more frequent term is higher: we reward the fact that the PMI is high despite its prevalence; this is less likely to be an artifact of chance.

Note that the rearranged expression seen in the second line of Equation \ref{eqn:apmi} is reminiscent of $\ppmi^{0.75}$ from \newcite{levy2015improving}.

The second log term in APMI is always negative, and we thus shift all values by a constant $k$ (chosen based on practical considerations of data size: the smaller the $k$, the larger the size of the sparse matrices; based on experimenting with various values of $k$, it appears that expansion quality is not very sensitive to $k$). Clipping this shifted value at 0 produces Asymmetrical PPMI (APPMI):
\begin{equation}
\appmi(w, c) = \max(0, \apmi(w, c) + k)
\end{equation}

The two matrices thus produced are shown in Equation \ref{eqn:matrices}. If we use PPMI instead of APPMI, these are transposes of each other.
\begin{equation}\label{eqn:matrices}
\begin{split}
\mitof_{w,c} &=\appmi(w, c)\\
\mftoi_{c, w} &= \appmi(c, w)    
\end{split}
\end{equation}

\subsection{Focused Similarity and Set Expansion}

We now come to the central idea of this paper: the notion of focused similarity. Typically, similarity is based on the dot product or cosine similarity of the context vectors. The pairwise similarity among all terms can be expressed as a matrix multiplication as shown in Equation \ref{eqn:mult}. Note that if we had used PPMI in Equation \ref{eqn:matrices}, the matrices would be each other's transposes and each entry in $\simmatrix$ in Equation \ref{eqn:mult} would be the dot-product-based similarity for a word pair.
\begin{equation}\label{eqn:mult}
\simmatrix=\mftoi\mitof
\end{equation}

We introduce context weighting by inserting a square matrix $W$ between the two (see Equation \ref{eqn:weighted_mult}). Similarity is unchanged if $W$ is the identity matrix. If $W$ is a non-identity diagonal matrix, this is equivalent to treating some contexts as more important than others. It is by appropriately choosing weights in $W$ that we achieve the context dependent similarity. If, for instance, all contexts other than those indicative of cars are zeroed out in $W$, \term{ford} and \term{obama} will have no similarity.

\begin{equation}\label{eqn:weighted_mult}
\simmatrix=\mftoi W\mitof
\end{equation}

\subsection{Set Expansion via Matrix Multiplication}

To expand a set of $k$ seeds, we can construct the $k$-hot column vector $S$ with a 1 corresponding to each seed, and a 0 elsewhere. Given $S$, we calculate the focus matrix, $W_S$. Then the expansion $E$ is a column vector that is just:
\begin{equation}\label{eqn:expansion}
E = \mftoi W_S \mitof S
\end{equation}

The score for a term in $E$ is the sum of its focused similarity to each seed.

\subsection{Motivating Our Choice of W}

When expanding the set \termset{taurus, cancer}---the set of star signs, or perhaps the constellations---we are faced with the presence of a polysemous term with a lopsided polysemy. The \term{disease} sense is much more prevalent than the \term{star sign} sense for \term{cancer}, and the associated contexts are also unevenly distributed. If we attempt to use Equation \ref{eqn:expansion} with the identity matrix $W$, the expansion is dominated by diseases.

The contexts we care about are those that are shared. Note that restricting ourselves to the intersection is not sensible, since if we are given a dozen seeds it is entirely possible that they share family resemblances and have a  high pairwise overlap in contexts between any two seeds but where there are almost no contexts shared by all. We thus require a soft intersection, and this we achieve by downweighting contexts based on what fraction of the seeds are associated with that context. The parameter $\rho$ described in the next section achieves this.

This modification helps, but it is not enough. Each disease-related context for \term{cancer} is now weakened, but their large number causes many diseases to rank high in the expansion. To address this, we can limit ourselves to only the top $n$ contexts (typically, $n=100$ is used). This way, if the joint contexts are highly ranked, the expansion will be based only on such contexts.

The \termset{taurus, cancer} example is useful to point out the benefits of an asymmetric association measure. Given \term{cancer}, the notion of \term{star sign} is not highly activated, and rightly so. If $w$ is \term{cancer} and $c$ is \featurename{Born Under X}, then $\ppmi(w, c)$ is low (as is $\appmi(w, c)$). However, $\appmi(c, w)$ is quite high, allowing us to highly score \term{cancer} when expanding \termset{taurus, aries}.

\subsection{Details of Calculating $W$}
 To produce $W$, we provide the seeds and two parameters: $\rho\in\mathbb{R}$ (the \emph{limited support penalty}) and $n\in\mathbb{N}$ (the \emph{context footprint}). Algorithm \ref{algo} provides the pseudo-code.
 
 First, we score contexts by their \term{activation} (line 3). We penalize contexts that are not supported by all the seeds: we produce the score by multiplying activation by $f^\rho$, where $f$ is the fraction of the seeds supporting that context (lines 5 and 7). Only the $n$ top scoring contexts will have non-zero values in $W$, and these get the value $f^\rho$.
 
\IncMargin{1em}
\begin{algorithm}[t]
  \SetKwInOut{Input}{input}\SetKwInOut{Output}{output}
  \Input{$S\subset \mathcal{V}$ (seeds), $\rho\in\R$ (limited support penalty), $n\in\mathbb{N}$ (context footprint)}
  \Output{The diagonal matrix $W$.}
  
  \BlankLine
  \For{$c\in \mathcal{C}$}{
    \emph{// Activation of the context.}
    
    $a(c) \leftarrow \sum_{w\in S}\mitof_{w, c}$
    
    \emph{// Fraction of $S$ with context active}
    
    $f(c) \leftarrow$ fraction with $\mitof_{*, c}>0$
    
    \emph{// Score of context}
    
 	$s(c) \leftarrow f(c)^\rho a(c)$
  }
  Sort contexts by score $s(c)$
  
  \For{$c\in C$}{
   \If{$c$ one of $n$ top-scoring contexts}{
   $W_{c, c} = f(c)^\rho$ }
  }
\caption{Calculating context focus}\label{algo}
\end{algorithm}

This notion of weighting contexts is similar to that used in the SetExpan framework \cite{shen2017setexpan}, although the way they use it is different (they use weighted Jaccard similarity based on context weights). Their algorithm for calculating context weights is a special case of our algorithm, with no notion of \emph{limited support penalty}, that is, they use $\rho=0$.

\subsection{Sparse Representations for Analogies}\label{analogy_matrix}

To solve the analogy problem ``What is the Ganga of Egypt?" we are looking for something that is like \term{Ganga} (this we can obtain via the set expansion of the (singleton) set \termset{Ganga}, as described above) and that we see often with \term{Egypt}, or to use Turney's terminology, in the same domain as \term{Egypt}.

To find terms that are in the same domain as a given term, we use the same statistical tools, merely with a different set of contexts. The context for a term is other terms in the same sentence. With this alternate definition of context, we produce $\dftoi$ exactly analogous to $\mftoi$ from Equation \ref{eqn:matrices}.

However, if we define $\ditof$ analogous to $\mitof$ and use these matrices for expansion, we run into unintended consequences since expanding \termset{evolution} provides not what things \term{evolution} is seen with, but rather those things that cooccur with what \term{evolution} co-occurs with. Since, for example, both \term{evolution} and \term{number} co-occur with \term{theory}, the two would appear related. To get around this, we zero out most non-diagonal entries in $\ditof$. The only off diagonal entries that we do not zero out are those corresponding to word pairs that seem to share a lemma (which we heuristically define as ``share more than 80\% of the prefix". Future work will explore using lemmas). An example of a pair we retain is \term{india} and \term{indian}), since when we are looking for items that co-occur with \term{india} we actually want those that occur with related words forms. An illustration for why this matters: \term{India} and \term{Rupee} occur together rarely (with a negative PMI) whereas \term{Indian} and \term{Rupee} have a strong positive PMI.

\subsection{Finding Analogies}

To answer ``What is the Ganga of Egypt", we use Equation \ref{eqn:expansion} on the singleton set \termset{ganga}, and the same equation (but with $\ditof$ and $\dftoi$) on \termset{egypt}. We intersect the two lists by combining the score of the shared terms in squash space (i.e., if the two scores are $m$ and $d$, the combined score is
\begin{equation}\label{eqn:joint}
\frac{100m}{99+m}+\frac{100d}{99+d}
\end{equation}

\section{Set Expansion Experiments and Evaluation}\label{sec:set_expansion}

\subsection{Experimental Setup}\label{section:expansion_expt}

We report data on two different corpora.  

\textbf{The Comparison Corpus.} We begin with 20 million English web pages randomly sampled from a set of popular web pages (high pagerank according to Google). We run Word2Vec on the text of these pages, producing a 200 dimensional embeddings. We also produce $\mitof$ and $\mftoi$ according to Equation \ref{eqn:matrices}. We use this corpus to compare Category Builder with Word2Vec-based techniques. Note that these web-pages may be noisier than Wikipedia. Word2Vec was chosen because it was deemed ``comparable": mathematically, it is an implicit factorization of the PMI matrix \cite{levy2014neural}.

\textbf{Release Corpus.} We also ran Category Builder on a much larger corpus. The generated matrices are restricted to the most common words and phrases (around 200,000). The matrices and associated code are publicly available\footnote{https://github.com/google/categorybuilder}.

\textbf{Using Word2Vec for Set Expansion.} Two classes of techniques are considered, representing members of both families described by \newcite{gyllensten2018distributional}. The centroid method finds the centroid of the seeds and expands to its neighbors based on cosine similarity. The other methods first find similarity of a candidate to each seed, and combines these scores using arithmetic, geometric, and harmonic means.

\textbf{Mean Average Precision (MAP).} MAP combines both precision and recall into a single number. The gold data to evaluate against is presented as sets of synsets, e.g., $\{\{\term{California}, \term{CA}\}, \{\term{Indiana}, \term{IN}\}, \ldots\}$.

An expansion $L$ consists of an ordered list of terms (which may include the seeds). Define $\precision_i(L)$ to be the fraction of items in the first $i$ items in $L$ that belong to at least one golden synset. We can also speak of the precision at a synset, $\precision_S(L) = \precision_j(L)$, where $j$ is the smallest index where an element in $S$ was seen in $L$. If no element in the synset $S$ was ever seen, then $\precision_S=0$. $\map(L)=avg(\precision_S(L))$ is the average precision over all synsets.

\textbf{Generalizations of MAP.} While $\map$ is an excellent choice for closed sets (such as \catname{U.S. states}), it is less applicable to open sets (say, \catname{Political Ideologies} or \catname{Scientists}). For such cases, we propose a generalization of $\map$ that preserves its attractive properties of combining precision and recall while accounting for variant names. The proposed score is $\map_n(L)$, which is the average of precision for the first $n$ synsets seen. That it is a strict generalization of $\map$ can be seen by observing that in the case of \catname{US States}, $\map(L)\equiv\map_{50}(L)$.

\subsection{Evaluation Sets}
We produced three evaluation sets, two closed and one open. For closed sets, following \newcite{wang2007language}, we use US States and National Football League teams. To increase the difficulty, for NFL teams, we do not use as seeds dismabiguated names such as \term{Detroit Lions} or \term{Green Bay Packers}, instead using the polysemous \term{lions} and \term{packers}. The synsets were produced by adding all variant names for the teams. For example, \term{Atlanta Falcons} are also known as \term{falcs}, and so this was added to the synset. 

For the open set, we use verbs that indicate things breaking or failing in some way. We chose ten popular instances (e.g., \term{break, chip, shatter}) and these act as seeds. We expanded the set by manual evaluation: any correct item produced by any of the evaluated systems was added to the list. There is an element of subjectivity here, and we therefore provide the lists used (Appendix \ref{appendix:lists}).

\subsection{Evaluation}

For each evaluation set, we did 50 set expansions, each starting with three randomly selected seeds.

\begin{table}[t]
\centering\small
\begin{tabular}{@{} l   c c c  @{}}
\multirow{2}{*}{\textbf{Technique}}&US&NFL&\emph{Break}\\
&States&Teams&Verbs\\\hline
W2V HM&            .858  &  .528       & .231 \\
W2V GM&            .864  &  .589       & .273 \\
W2V AM&            .852  &  .653       & .332 \\
W2V Centroid&          .851  &  .646       & .337 \\\hline
CB:$\ppmi;\rho=0$&    .918  &  .473       & .248\\
CB:$\ppmi;\rho=3$&    \textbf{.922}  &  .612       & .393 \\
CB:$\appmi;\rho=0$&   .900  &  .584       & .402 \\
CB:$\appmi;\rho=3$&   .907  &  \textbf{.735}       & \textbf{.499} \\\hline
CB:Release Data$^\dagger$&      .959  &  .999      & .797 \\\hline
\end{tabular}
\caption{MAP scores on three categories. The first four rows use various techniques with Word2Vec. The next four demonstrate Category Builder built on the same corpus, to show the effect of $\rho$ and association measure used. For all four Category Builder rows, we used $n=100$. Both increasing $\rho$ and switching to APPMI can be seen to be individually and jointly beneficial.  $^\dagger$The last line reports the score on a different corpus, the release data, with APPMI and $\rho=3, n=100$.}\label{table:expansion}
\end{table}

\begin{table}[hbtp]
\centering\small
\begin{tabular}{@{} p{1.6cm} | c c c c c c @{}}
&5&10&30&50&100&500\\\hline
US States&\textbf{.932}&.925&.907&.909&.907&.903\\
NFL& .699 & .726 & .731 & .734 & \textbf{.735} & .733\\
Break Verbs&.339&.407&.477&.485&.496&\textbf{.511}\\\hline
\end{tabular}
\caption{Effect of varying n. APPMI with $\rho=3$.}\label{table:size}
\end{table}

\textbf{Effect of $\rho$ and APPMI.} Table \ref{table:expansion} reveals that APPMI performs better than PPMI --- significantly better on two sets, and slightly worse on one. Penalizing contexts that are not shared by most seeds (i.e., using $\rho>0$) also has a marked positive effect. 

\textbf{Effect of $n$.} Table \ref{table:size} reveals a curious effect. As we increase $n$, for \catname{US States}, performance drops somewhat but for \catname{Break Verbs} it improves quite a bit. 
Our analysis shows that pinning down what a state is can be done with very few contexts, and other shared contexts (such as \featurename{live in X}) are shared also with semantically related entities such as states in other countries. At the other end, \catname{Break Verbs} is based on a large number of shared contexts and using more contexts is beneficial.  

\subsection{Error Analysis.} Table \ref{table:intrusion} shows the top errors in expansion. The kinds of drifts seen in the two cases are revealing. Category Builder picks up word fragments (e.g., because of the US State \term{New Mexico}, it expanded states to include \term{Mexico}). It sometimes expands to a hypernym (e.g., \term{province}) or siblings (e.g., instead of Football teams sometimes it got other sport teams). With Word2Vec, we see similar errors (such as expanding to the semantically similar \term{southern california}).

\begin{table}[t]
\centering\small
\begin{tabular}{@{} p{1.2cm}  p{0.8cm} | p{4.4cm}@{}}
Set&Method&Top Errors\\\hline
\multirow{2}{*}{US States}&W2V&southern california; east tennessee; seattle washington \\
&CB&carolina; hampshire; dakota; ontario; jersey; province\\\hline
\multirow{2}{*}{NFL}&W2V&hawks; pelicans; tigers; nfl; quarterbacks; sooners\\
&CB&yankees; sox; braves; mets; knicks; rangers, lakers\\\hline
\end{tabular}
\caption{Error analysis for US States and NFL. Arithmetic Mean method is used for W2V and $\rho=3$ and APPMI for Category Builder}\label{table:intrusion}
\end{table}

\subsection{Qualitative Demonstration}
Table \ref{table:detailed} shows a few examples of expanding categories, with $\rho=3, n=100$. 

Table \ref{table:synthetic} illustrates the power of Category Builder by considering a a synthetic corpus produced by replacing all instances of \term{cat} and \term{denver} into the hypothetical \term{CatDenver}. This illustrates that even without explicit WSD (that is, separating \term{CatDenver} to its two ``senses", we are able to expand correctly given an appropriate context. 
To complete the picture, we note that expanding \termset{kitten, dog} as well as \termset{atlanta, phoenix} contains \term{CatDenver}, as expected.

\begin{table}[hbtp]
\centering\small
\begin{tabular}{@{} p{1.2cm} p{5.8cm}  @{}}
\textbf{Seeds}&\textbf{CB Expansion, $\rho=3, n=100$} \\
\hline
ford, nixon&nixon, ford, obama, clinton, bush, richard nixon, reagan, roosevelt, barack obama, bill clinton, ronald reagan, w. bush, eisenhower\\\hline
ford,\hspace{.5cm}chevy&ford, chevy, chevrolet, toyota, honda, nissan, bmw, hyundai, volkswagen, audi, chrysler, mazda, volvo, gm, kia, subaru, cadillac\\\hline
ford,\hspace{.5cm}depp&ford, depp, johnny depp, harrison ford, dicaprio, tom cruise, pitt, khan, brad pitt, hanks, tom hanks, leonardo dicaprio\\\hline
safari, trip$^\dagger$&trip, safari, tour, trips, cruise, adventure, excursion, vacation, holiday, road trip, expedition, trek, tours, safaris, journey, \\\hline
safari,\hspace{.5cm}ie$^\dagger$&safari, ie, firefox, internet explorer, chrome, explorer, browsers, google chrome, web browser, browser, mozilla firefox\\\hline
\end{tabular}
\caption{Expansion examples using Category Builder so as to illustrate its ability to deal with Polysemy. $^\dagger$ For these examples, $\rho=5$ }\label{table:detailed}
\end{table}

\begin{table}[t]
\centering\small
\begin{tabular}{@{} p{1.8cm} p{5.2cm}  @{}}
\textbf{Seeds}&\textbf{CB Expansion, $\rho=3, n=100$} \\
\hline
CatDenver, dog&dogs, cats, puppy, pet, rabbit, kitten, animal, animals, pup, pets, puppies, horse\\\hline
CatDenver, phoenix&chicago, atlanta, seattle, dallas, boston, portland, angeles, los angeles\\\hline
CatDenver, TigerAndroid&cats, lion, dog, tigers, kitten, animal, dragon, wolf, dogs, bear, leopard, rabbit\\\hline
\end{tabular}
\caption{Expansion examples with synthetic polysemy by replacing all instances of \term{cat} and \term{denver} into the hypothetical \term{CatDenver} (similarly, \term{TigerAndroid}). A single other term is enough to pick out the right sense.}\label{table:synthetic}
\end{table}

\section{Analogies}
\subsection{Experimental Setup}
We evaluated the analogy examples used by \newcite{mikolov2013linguistic}. Category Builder evaluation were done by expanding using syntactic and sentence-based-cooccurrence contexts as detailed in Section \ref{analogy_matrix} and scoring items according to Equation \ref{eqn:joint}. For evaluating using Word2Vec, the standard vector arithmetic was used.

In both cases, the input terms from the problem were removed from candidate answers (as was done in the original paper). \newcite{linzen2016issues} provides analysis and rationales for why this is done.

\subsection{Evaluation}

Table \ref{table:analogies} provides the evaluations. A few words are in order for the difference between the published scores for Word2Vec analogies elsewhere (e.g., \citealp{linzen2016issues}). Their reported numbers for common capitals were around 91\%, as opposed to around 87\% here. Where as Wikipedia is typically used as a corpus, that was not the case here. Our corpus is noisier, and may not have the same level of country-based factual coverage as Wikipedia, and almost all non-grammar based analogy problems are of that nature.

A second matter to point out is why grammar based rows are missing from Table \ref{table:analogies}. Grammar based analogy classes cannot be solved with just two terms. For \analogy{boy}{boys}{king}{?}, dropping the first term \term{boy} takes away information crucial to the solution in a way that dropping the first term of \analogy{US}{dollar}{India}{?} does not. The same is true for the \term{family} class of analogies.

\begin{table}[t]
\centering\small
\begin{tabular}{@{}  p{2.3cm} | c | c c  @{}}
&\analogy{a}{b}{c}{?}&\multicolumn{2}{c}{Harder \analogy{}{b}{c}{?}(\term{a} withheld)}\\
\hspace*{0pt}\hfill\textbf{Method}&\textbf{W2V}&\multicolumn{2}{c}{\textbf{CB:APPMI}}\\\hline
\hspace*{0pt}\hfill\textbf{Corpus}&Comp&Comp&\textbf{Release$^\dagger$}\\\hline
common capitals    &.872  &\textbf{.957} & .941\\
city-in-state      &.657  &\textbf{.972} & .955\\
currency           &.030  &\textbf{.037} & .122\\
nationality        &.515  &\textbf{.615} & .655\\
world capitals     &.472  &\textbf{.789} & .668\\
family             &\textbf{.617}  &.217 & .306\\
\end{tabular}
\caption{Performance on Analogy classes from \newcite{mikolov2013linguistic}. The first two columns are derived from the same corpus, whereas the last column reports numbers on the data we will release. For category builder, we used $\rho=3, n=100$}\label{table:analogies}
\end{table}

\subsection{Qualitative Demonstration}

Table \ref{table:analogy_sampler} provides a sampler of analogies solved using Category Builder. 

\begin{table}[t]
\centering\small
\begin{tabular}{@{}c c c @{}}
$term_1$&$term_2$&\textbf{What is the $term_1$ of $term_2$?}\\
\hline
voldemort&tolkien&sauron\\
voldemort&star wars&vader\\
ganga&egypt&nile\\
dollar&india&rupee\\
football&india&cricket\\
civic&toyota&corolla\\
\end{tabular}
\caption{A sampler of analogies solved by Category Builder.}\label{table:analogy_sampler}
\end{table}

\section{Limitations}

Much work remains, of course. The analogy work presented here (and also the corresponding work using vector offsets) is no match for the subtlety that people can bring to bear when they see deep connections via analogy. Some progress here could come from the ability to discover and use more semantically meaningful contexts.

There is currently no mechanism to automatically choose $n$ and $\rho$. Standard settings of $n=100$ and $\rho=3$ work well for the many applications we use it for, but clearly there are categories that benefit from very small $n$ (such as \catname{Blue Things}) or very large $n$. Similarly, as can be seen in Equation \ref{eqn:joint}, analogy also uses a parameter for combining the results, with no automated way yet to choose it. Future work will prioritize this.

The current work suggests, we believe, that it is beneficial to not collapse the large dimensional sparse vector space that implicitly underlies many embeddings. Having the ability to separately manipulate contexts can help differentiate between items that differ on that context.  That said, the smoothing and generalization that dimensionality reduction provides has its uses, so finding a combined solution might be best. 

\section{Conclusions}
Given that natural categories vary in their degree of similarities and their kinds of coherence, we believe that solutions that can adapt to these would perform better than context independent notions of similarity.

As we have shown, Category Builder displays the ability to implicitly deal with polysemy and determine similarity in a context sensitive manner, as exhibited in its ability to expand a set by latching on to what is common among the seeds. 

In developing it we proposed a new measure of association between words and contexts and demonstrated its utility in set expansion and a hard version of 
the analogy problem.  
In particular, our results show that sparse representations deserve additional careful study.

\section*{Acknowledgments}

We are thankful to all the reviewers for their helpful comments and critiques. In particular, Ido Dagan, Yoav Goldberg, Omer Levy, Praveen Paritosh, and Chris Waterson gave us insightful comments on earlier versions of this write up. The research of Dan Roth is partly supported by a Google gift and by DARPA, under agreement number FA8750-13-2-008.  

\bibliography{acl2018}
\bibliographystyle{acl_natbib}

\appendix

\section{Supplemental Material}

\subsection{Lists Used in Evaluating Set Expansion}\label{appendix:lists}

\textbf{US States.} Any of the 50 states could be used as a seed. The 50 golden synsets were the 50 pairs of state name and abbreviation (e.g., \termset{California, CA}).

\textbf{NFL Teams.} Any of the first terms in these 32 synsets could be used as a seed. The golden synsets are: 
\termset{Bills, Buffalo Bills}, 
\termset{Dolphins, Miami Dolphins, Phins},
\termset{Patriots, New England Patriots, Pats},
\termset{Jets, New York Jets},
\termset{Ravens, Baltimore Ravens},
\termset{Bengals, Cincinnati Bengals},
\termset{Browns, Cleveland Browns},
\termset{Steelers, Pittsburgh Steelers},
\termset{Texans, Houston Texans},
\termset{Colts, Indianapolis Colts},
\termset{Jaguars, Jacksonville Jaguars, Jags},
\termset{Titans, Tennessee Titans},
\termset{Broncos, Denver Broncos},
\termset{Chiefs, Kansas City Chiefs},
\termset{Chargers, Los Angeles Chargers},
\termset{Raiders, Oakland Raiders},
\termset{Cowboys, Dallas Cowboys},
\termset{Giants, New York Giants},
\termset{Eagles, Philadelphia Eagles},
\termset{Redskins, Washington Redskins},
\termset{Bears, Chicago Bears},
\termset{Lions, Detroit Lions},
\termset{Packers, Green Bay Packers},
\termset{Vikings, Minnesota Vikings, Vikes},
\termset{Falcons, Atlanta Falcons, Falcs},
\termset{Panthers, Carolina Panthers},
\termset{Saints, New Orleans Saints},
\termset{Buccaneers, Tampa Bay Buccaneers, Bucs},
\termset{Cardinals, Arizona Cardinals},
\termset{Rams, Los Angeles Rams},
\termset{49ers, San Francisco 49ers, Niners}, and
\termset{Seahawks, Seattle Seahawks}

\textbf{Break Verbs.} Seeds are chosen from among these ten items: \emph{break, chip, shatter, rot, melt, scratch, crush, smash, rip, fade}. Evaluation is done for $\map_{30}$ (see Section \ref{section:expansion_expt}). The following items are accepted in the expansion: \emph{break up, break down, tip over, splinter, tear, come off, crack, disintegrate, deform, crumble, burn, dissolve, bend, chop, stain, destroy, smudge, tarnish, explode, derail, deflate, corrode, trample, ruin, suffocate, obliterate, topple, scorch, crumple, pulverize, fall off, cut,
dry out, split, deteriorate, hit, blow, damage, wear out, peel, warp, shrink, evaporate, implode, scrape, sink, harden, abrade, unhinge, erode,
calcify, vaporize, sag, shred, degrade, collapse, annihilate}. In the synsets, we also added the morphological variants (e.g., \termset{break, breaking, broke, breaks}).

\subsection{Word2Vec Model Details}
The word2vec model on the ``comparison corpus" created 200 dimensional word embeddings. We used a skip-gram model with a batch size of 100, a vocabulary of 600k ngrams, and negative sampling with 100 examples. It was trained using a learning rate of 0.2 with Adagrad optimizer for 70 million steps.

\end{document}